\documentclass{article}
\usepackage[english]{babel}
\usepackage[T1]{fontenc}
\usepackage[letterpaper,top=2cm,bottom=2cm,left=3cm,right=3cm,marginparwidth=1.75cm]{geometry}

\usepackage{amsmath,amsfonts,amssymb}
\usepackage{graphicx}
\usepackage[colorlinks=true, allcolors=blue]{hyperref}

\title{Modern graph neural networks do worse than classical greedy algorithms in solving combinatorial optimization problems like Maximum Independent Set}
\author{Maria Chiara Angelini$^{1,2}$ \and Federico Ricci-Tersenghi$^{1,2,3}$}
\date{\footnotesize $^1$Dipartimento di Fisica, Sapienza Universit\`a di Roma, P.le Aldo Moro 5, 00185 Rome, Italy\\
$^2$Istituto Nazionale di Fisica Nucleare, Sezione di Roma I, P.le A. Moro 5, 00185 Rome, Italy\\
$^3$Institute of Nanotechnology (NANOTEC) - CNR, Rome unit, P.le A. Moro 5, 00185 Rome, Italy}

\begin{document}
\maketitle

Author to whom correspondence should be addressed: Maria Chiara Angelini, e-mail address maria.chiara.angelini@roma1.infn.it

\section{Introduction}
Arising from Martin J. A.
Schuetz et al. Nature Machine Intelligence
https://doi.org/10.1038/s42256-022-00468-6 (2022).

The recent work ``Combinatorial Optimization with Physics-Inspired Graph Neural Networks'' [Nat Mach Intell 4 (2022) 367] introduces a physics-inspired unsupervised Graph Neural Network (GNN) to solve combinatorial optimization problems on sparse graphs. To test the performances of these GNNs, the authors of the work show numerical results for two fundamental problems: maximum cut and maximum independent set (MIS), concluding \guillemotleft that the graph neural network optimizer performs on par or outperforms existing solvers, with the ability to scale beyond the state of the art to problems with millions of variables.\guillemotright
In this comment, we show that a simple greedy algorithm, running in almost linear time, can find solutions for the MIS problem of much better quality than the GNN in a much shorter time. In general, many claims of superiority of neural networks in solving combinatorial problems are at risk of being not solid enough, since we lack standard benchmarks based on really hard problems. We propose one of such hard benchmarks, and we hope to see future neural network optimizers tested on these problems before any claim of superiority is made.

The last years have seen an incredible increase in the use of neural networks to solve all kinds of problems, both in applications and in fundamental science;
Discrete combinatorial optimization problems make no exception \cite{cappart2021combinatorial, kotary2021end, bengio2021machine} and are extremely relevant, given that they are often at the basis of our understanding of the fundamental computational limits.

For these reasons the proposal made in Ref.~\cite{schuetz2022combinatorial} to use physics-inspired unsupervised Graph Neural Network (GNN) to solve combinatorial optimization problems defined on a graph seemed very promising and got published in a journal with high impact.
The authors of Ref.~\cite{schuetz2022combinatorial} tested the performance of the GNN on two standard optimization problems: the maximum cut and maximum independent set (MIS). 
A very good property of this newly introduced GNN optimizer is that it can scale to problem instances much larger than what many previous deep-learning approaches could handle \cite{Selsam2019,cameron2020predicting,toenshoff2021graph}.

Let us focus on the MIS problem, which is defined as follows. Given an undirected random regular graph of fixed degree $d$ ($d$-RRG) with $n$ nodes, an independent set (IS) is a subset of vertices not containing any pair of nearest neighbors. The MIS problem then requires finding the largest IS, whose size is called $\alpha$. The distribution of the MIS density $\alpha/n$ among different $d$-RRG concentrates for large $n$ to a value $\rho(d)$ which depends just on the degree $d$. The MIS is an NP-hard problem, however, one can hope to find an algorithm that finds in polynomial time an IS whose size is as close as possible to the maximal one. Moreover, the performances of a good algorithm should not degrade for larger $n$.

\begin{figure}
\centering
\includegraphics[width=\textwidth]{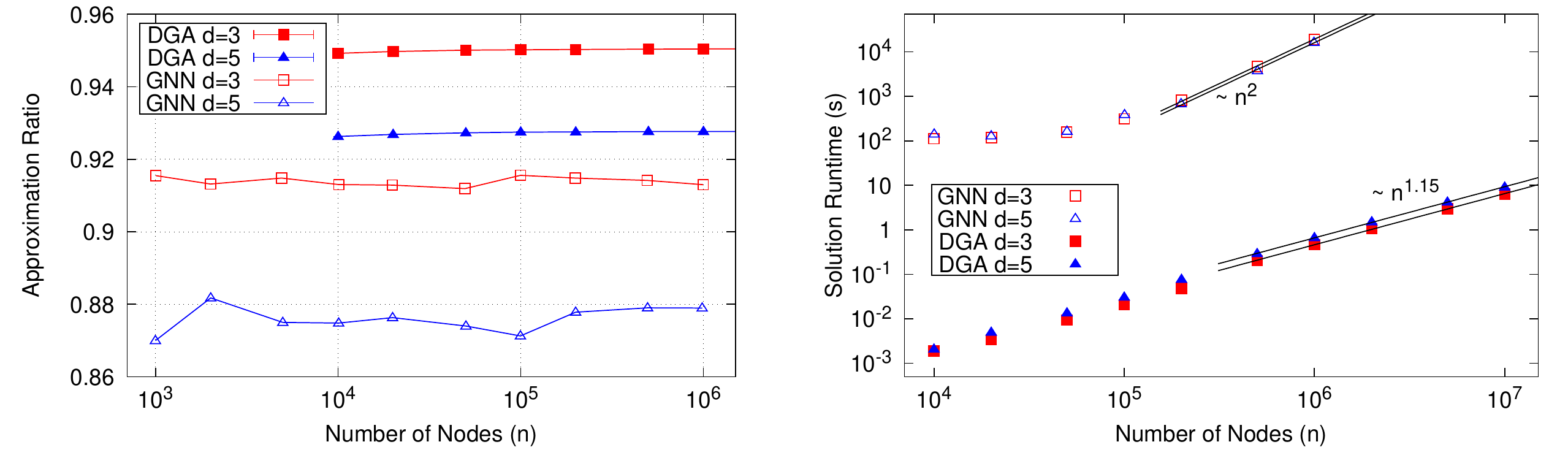}
\caption{\label{fig:times} Comparison of the Graph Neural Network (GNN) of Ref.~\cite{schuetz2022combinatorial} and a simple degree-based greedy algorithm (DGA) in computing Independent Sets (IS) in $d$-regular random graphs.
\textbf{Left panel}: The size of the IS found by DGA (full symbols) and by GNN (empty symbols), reported as an approximation ratio with respect to the known theoretical upper bounds. DGA finds much better solutions than GNN.
\textbf{Right panel}: Running time in seconds for DGA (full symbols) and GNN (empty symbols). The DGA scaling is much better than the one for GNN. At $n=10^6$ the speedup of DGA is larger than a factor $10^4$ with respect to GNN.}
\end{figure}

The GNN of Ref.~\cite{schuetz2022combinatorial} can find IS for graphs of very large sizes ($n\le 10^6$): the run-time is proportional to a small power of the problem size, $t \sim n^{1.7}$, and the performances are stable with $n$ (this is highly non-trivial, looking at previous NN approaches to optimization problems \cite{Selsam2019,cameron2020predicting,toenshoff2021graph}).
The size of the IS found by GNN and the running times are reported with open symbols in Fig.~\ref{fig:times}.

Problems in Ref.~\cite{schuetz2022combinatorial} arise when comparing the GNN performances with other available algorithms and thus claiming the superiority of the approach based on GNN. The authors of Ref.~\cite{schuetz2022combinatorial} consider only the Boppana-Halldorsson (BH) approximated algorithm \cite{boppana1992approximating} that shows a run time scaling as $t \sim n^{2.9}$ in the range $n\le500$.

\section{Comparison with simple greedy algorithms} 

However, there exist many other algorithms for computing IS, that work much faster than BH and the new GNN optimizer should be compared with these too.

The simplest possible algorithm that one can design to solve the MIS problem is the Greedy Algorithm (GA) \cite{karp1981maximum} that takes a time to reach a solution that is linear in the problem size $n$.
GA works as follows.
It starts from an empty IS and $G^0$ being the original graph.
At each step $t$, a node is chosen at random from the graph $G^t$ and added to the IS.
All its neighbors are removed from the graph $G^t$, thus getting a new graph $G^{t+1}$. 
At the time $t^*$ such that $G^{t^*}$ is empty, the process stops.

One can also construct an improved version of the GA by exploiting the degrees of the nodes. In the degree-based GA (DGA), at each step, one chooses the node with the smallest degree in $G^t$ \cite{wormald1995differential}.
This algorithm runs in a time almost linear in $n$ if the degree-based ordering of the nodes is managed smartly.

In Fig.~\ref{fig:times} we compare the performances of the GNN of Ref.~\cite{schuetz2022combinatorial} (empty symbols) and DGA (full symbols) in finding MIS on $d$-RRG with $d=3,5$.
In the left panel we report the quality of the solutions found by the two algorithms, shown as the Approximation Ratio (AR) to the current best upper bounds $\rho_\text{\tiny UB}(d=3) = 0.45537$ and $\rho_\text{\tiny UB}(d=5) = 0.38443$ \cite{mckay1987independent}. These bounds are likely to be not strict as the replica method provides an optimal AR smaller than 1, namely $\mathsf{AR}_\text{\tiny 1RSB}(d=3)\simeq 0.990$ and $\mathsf{AR}_\text{\tiny 1RSB}(d=5)\simeq 0.987$ \cite{barbier2013hard}.
DGA clearly outperforms the GNN of ref. \cite{schuetz2022combinatorial}, especially in the case $d=5$, where the IS found by DGA are 6\% larger than those found by that GNN.

In the right panel, we show the running times for both algorithms: data for DGA have been collected by running on a 2.3 GHz MacBook Pro while data for the GNN have been extracted from Fig.~5 in Ref.~\cite{schuetz2022combinatorial}. The latter correspond to the aggregated run-time that includes the post-processing, because the authors furnish no additional information on the time needed for the different steps of the computation. However, the post-processing time to check if the output configuration is an IS should be linear in $n$, so most of the time should be dedicated to the GNN computation.
The scaling of the running times with the problem size is much better for DGA than for GNN, being the former almost linear in $n$ (the exponent 1.15 is probably due to pre-asymptotic effects), while the last data points for GNN scale close to quadratically in $n$. 
Not only the scaling is better for DGA, but the actual running times are orders of magnitude faster. For example, for $n=10^6$ DGA shows a speedup with respect to GNN by a factor larger than $10^4$.

We thus think that the claim in Ref.~\cite{schuetz2022combinatorial} \guillemotleft\textit{We find that the graph neural network optimizer performs on par or outperforms existing solvers, with the ability to scale beyond the state of the art to problems with millions of variables}\guillemotright\ is not supported by the data shown here and should be modified.

\section{Discussion}

We have reported in detail the performances of DGA because we believe that such a simple greedy algorithm should be considered as a \emph{minimal benchmark} and any new algorithm must perform at least better than DGA to be taken into serious consideration.

But DGA is not the end of the story. There exist many other standard algorithms which do better than DGA.
A thorough study of the performances of these algorithms to solve the MIS problem can be found in Ref.~\cite{angelini2019monte}. For completeness, we just report here the AR achieved by some of them (addressing the reader to Ref.~\cite{angelini2019monte} for further details). Both Simulated Annealing and Parallel Tempering, two very effective algorithms based on MCMC, reach $\mathsf{AR}_\text{\tiny MCMC}(d=3)\simeq 0.984$ and $\mathsf{AR}_\text{\tiny MCMC}(d=5)\simeq 0.981$. While Belief Propagation with Reinforcement, a very effective message passing algorithm, can reach $\mathsf{AR}_\text{\tiny BPR}(d=3)\simeq 0.987$ and $\mathsf{AR}_\text{\tiny BPR}(d=5)\simeq 0.981$.

Not only the performances of these standard algorithms are overwhelming better than the GNN of Ref.~\cite{schuetz2022combinatorial}, but they approach closely the supposedly optimal AR computed via the replica method, $\mathsf{AR}_\text{\tiny 1RSB}(d=3)\simeq 0.990$ and $\mathsf{AR}_\text{\tiny 1RSB}(d=5)\simeq 0.987$ \cite{barbier2013hard}.
This observation suggests that finding the optimal MIS in $d$-RRG with $d=3,5$ is not a really hard problem and the optimum can be well approximated by algorithms running in polynomial time in $n$.
Indeed a statistical physics study investigating the structure of IS in $d$-RRG \cite{barbier2013hard} found that only for $d>16$, increasing the IS size, the space of IS undergoes a clustering transition, that is usually related to hardness in sampling.
For $d<16$ the structure of IS is such that the MIS is likely to be easy to approximate.

A fundamental question at the time of evaluating a new optimization algorithm is the following: \textit{``What are the really hard problems that should be used as a benchmark to test algorithmic performances?''}

We have argued that for the MIS using $d$-RRG with $d<16$ is likely to be an easy problem and the test would be not very selective. 
However, for larger $d$ we expect the optimization to become much more demanding because the clustering of the IS of large size is likely to create relevant barriers that affect any algorithm searching for the MIS.
This picture is supported by analytical results in the large $d$ limit, where no algorithm is known to find IS of density larger than $\rho_\text{alg}(d)=\log(d)/d$, even if the MIS is known to have density $\rho_\text{max}(d)=2\log(d)/d$ \cite{coja2015independent}: that is, no algorithm achieves an AR better than 0.5 in this limit.

So, a possible answer to the fundamental question above is to study MIS on $d$-RRG with $d>16$. And start by comparing with the results presented in Ref.~\cite{angelini2019monte} for $d=20$ and $d=100$.
Obviously, a good optimization algorithm should run in a time polynomial (better if linear) in $n$, and the quality of solutions found should be better than simple existing algorithms and should not degrade increasing $n$.

In our opinion, at present, optimizers based on neural networks (like the one presented in Ref.~\cite{schuetz2022combinatorial}) do not satisfy the above requirements and are not able to compete with simple standard algorithms to solve hard optimization problems. 
We showed that this is true for the GNN introduced in Ref.~\cite{schuetz2022combinatorial} applied to the MIS problem and 
the same conclusion holds also in the case of the Max-Cut problem on sparse graphs, as shown in the concurrent comment by Boettcher.

One could argue that both these examples are analyzing problems on sparse graphs, and NN can be more effective on denser graphs.
However, there are already results allowing for such a comparison in dense combinatorial problems: for example, in the problem of recovering a planted clique in a dense graph, the NN introduced in Ref.~\cite{levinas2022planted} do not reach the performances of message passing algorithms \cite{deshpande2015finding} or those of the Monte Carlo Parallel Tempering method \cite{angelini2018parallel}.

Concluding, we believe it is of primary importance to understand whether and when neural networks can become competitive in solving hard problems or whether there is any deeper reason for their failure.

The authors declare no competing interests.

\end{document}